\documentclass[submission,copyright,creativecommons]{eptcs}
\providecommand{\event}{ICLP 2021} 
\usepackage{underscore}           

\usepackage{soul}
\usepackage{url}
\usepackage[utf8]{inputenc}
\usepackage{amsmath}
\usepackage{amssymb}
\usepackage{booktabs}
\usepackage{algorithm}
\usepackage{algpseudocode}
\usepackage{listings}
\usepackage{tikz}
\usepackage{tikzscale}
\usetikzlibrary{arrows}
\usetikzlibrary{shapes.geometric}
\usepackage[switch]{lineno}
\usetikzlibrary{shapes.geometric}
\usetikzlibrary{arrows,backgrounds}
\usetikzlibrary{fit}
\usetikzlibrary{positioning}

\usepackage{fancyvrb}
\usepackage{mathptmx}

\title{Generating Explainable Rule Sets from Tree-Ensemble Learning Methods by Answer Set Programming}
\author{
Akihiro Takemura \qquad\qquad Katsumi Inoue
\institute{The Graduate University for Advanced Studies, SOKENDAI, Japan\\
National Institute of Informatics, Japan}
\email{atakemura@nii.ac.jp \;\qquad\qquad\; inoue@nii.ac.jp}
}

\def\titlerunning{Generating Explainable Rule Sets from Tree-Ensemble Learning Methods by ASP}
\def\authorrunning{A. Takemura \& K. Inoue}
\begin{document}
\maketitle

\begin{abstract}
We propose a method for generating explainable rule sets from tree-ensemble learners using Answer Set Programming (ASP). To this end, we adopt a decompositional approach where the split structures of the base decision trees are exploited in the construction of rules, which in turn are assessed using pattern mining methods encoded in ASP to extract interesting rules. We show how user-defined constraints and preferences can be represented declaratively in ASP to allow for transparent and flexible rule set generation, and how rules can be used as explanations to help the user better understand the models. Experimental evaluation with real-world datasets and popular tree-ensemble algorithms demonstrates that our approach is applicable to a wide range of classification tasks.
\end{abstract}

\section{Introduction}

Interpretability in machine learning is the ability to explain or to present in understandable terms to a human \cite{doshi-velezRigorousScienceInterpretable2017}. Interpretability is particularly important when, for example the goal of the user is to gain knowledge from some form of explanations about the data or process through machine learning models, or when making high-stakes decisions based on the outputs from the machine learning models where the user has to be able to trust the models.

In this work we address the problem of explaining and understanding tree-ensemble learners by extracting meaningful rules from them. This problem is of practical relevance in business domains where the understanding of the behavior of high-performing machine learning models and extraction of knowledge in human readable form can aid users in the decision making process. We use {\it Answer Set Programming (ASP)} \cite{gelfondStableModelSemantics1988,lifschitzWhatAnswerSet2008} to generate rule sets from tree-ensembles.
ASP is a declarative programming paradigm for solving difficult search problems. An advantage of using ASP is its expressiveness and extensibility, especially when representing constraints. To our knowledge, ASP has never been used in the context of rule sets generation from tree-ensembles, although it has been used in pattern mining, e.g.,  \cite{jarvisaloItemsetMiningChallenge2011a,guyetUsingAnswerSet,DBLP:conf/ijcai/GebserGQ0S16,paramonovHybridASPbasedApproach2019}.


Generating interpretations for machine learning models is a challenging task since it is often necessary to account for multiple competing objectives. For instance, if accuracy is the most important metric, then it is in direct conflict with interpretability, because accuracy favors specialization while interpretability favors generalization. Any interpretation method should also strive to imitate the behavior of learned models as to minimize misrepresentation of models, which in turn may result in misinterpretation by the user. While there are many interpretation methods available (some are covered in Section 2), we propose to use ASP as a medium to represent the user requirements declaratively and to quickly search feasible solutions for faster prototyping. By implementing a rule selection method as a post-processing step to model training, we aim to offer an off-the-shelf objective interpretation tool as an alternative to subjective manual rule selection, which could be applied to existing processes with minimum modification.

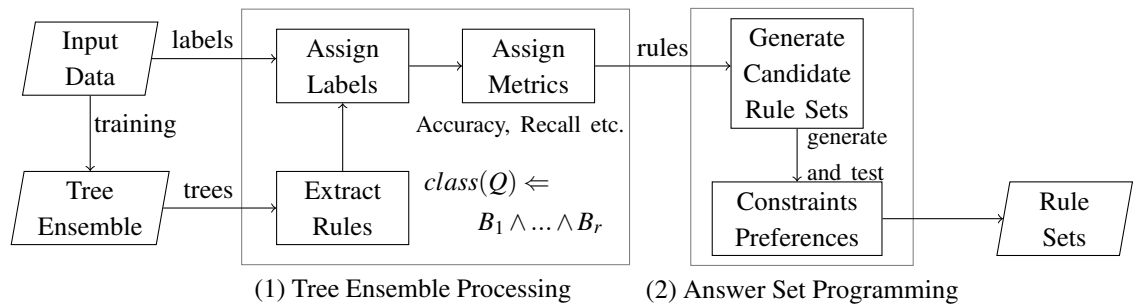
\begin{figure}[tb]
  \centering 
    \begin{tikzpicture}

\usetikzlibrary{shapes.geometric, arrows,backgrounds,fit,positioning}
\tikzset{
input/.style={trapezium, trapezium left angle=80, trapezium right angle=100, 
draw,  text centered, text width=1.2cm, minimum height=0.6cm},
input2/.style={trapezium, trapezium left angle=80, trapezium right angle=100,
draw,  text centered, text width=1.5cm, minimum height=0.6cm},
tp/.style={rectangle,  text centered, text width=7cm, minimum height=0.6cm},
tp1/.style={rectangle,  draw, text centered, text width=1.5cm, minimum height=0.6cm},
rule/.style={rectangle,  draw,  text centered, text width=7cm, minimum height=0.6cm},
asp/.style={rectangle,  draw,  text centered, text width=7cm, minimum height=0.6cm},
outputrule/.style={trapezium, trapezium left angle=60, trapezium right angle=120, draw,  text centered, text width=5cm, minimum height=0.6cm},
};

\node[input] (a1) {\small Input Data};
\node[input2, below=1cm of a1] (a2) {\small Tree Ensemble};
\draw[->] (a1) -- (a2) node[midway,above,xshift=.6cm,yshift=-.2cm]{\small training};

\node[tp1, right=1.5cm of a2,] (b1) {\small Extract Rules};
\node[right=.05cm of b1, yshift=.3cm, text width=2.4cm] (b1t) {\small \begin{align*}
    class&(Q) \Leftarrow\\ & B_1 \wedge ... \wedge  B_r
\end{align*}};
\node[tp1, right=1.5cm of a1.north, anchor=north, above=.9 of b1] (b2) {\small Assign Labels};
\node[tp1, right=.7cm of b2] (b3) {\small Assign Metrics};
\node[below=.03cm of b3, text width=3.cm] (b3t) {\footnotesize Accuracy, Recall etc.};
\node[fit=(b1) (b2) (b3), draw=gray, text width=4.9cm,  minimum height=3.4cm] (teep) {};
\node[below=.05cm of teep, xshift=-0.3cm] (b) {\small (1) Tree Ensemble Processing};

\draw[->] (a1.east) -- (a1.east -| b2.west) node[midway,above,xshift=-.15cm]{\small labels};
\draw[->] (a2) -- (b1) node[midway,above,xshift=-.15cm]{\small trees};
\draw[->] (b1) -- (b2);
\draw[->] (b2) -- (b3);

\node[tp1, right=1.8cm of b3, yshift=-.1cm] (c1) {\small Generate Candidate Rule Sets};
\node[tp1, below=.7cm of c1, text width=2.cm] (c2) {\small Constraints Preferences};
\node[fit=(c1) (c2), draw=gray, text width=2.7cm, minimum height=3.4cm, right=.8cm of teep] (aspe) {};
\node[below=.05cm of aspe] (c) {\small (2) Answer Set Programming};

\draw[->] (b3.east) -- (b3.east -| c1.west) node[midway,above,]{\small rules};
\draw[->] (c1.south) -- (c2.north) node[midway,right,text width=1.3cm]{\footnotesize generate and test};

\node[input, right=1.6cm of c2] (d1) {\small Rule Sets};
\draw[->] (c2) -- (d1);

\end{tikzpicture}
    \vspace{-.5\baselineskip}
  \caption{Overview of our framework}
  \label{fig:pipeline}
  \vspace{-.5\baselineskip}
\end{figure}

We consider the two-step procedure for rule set generation from tree-ensembles (Figure \ref{fig:pipeline}): (1) extracting rules from trained decision tree-ensembles, and (2) computing sets of rules according to selection criteria and preferences encoded declaratively in ASP. For the first step, we employ the efficiency and prediction capability of modern tree-ensemble algorithms in finding useful feature partitions for prediction from data. For the second step, we exploit the expressiveness of ASP in encoding constraints and preference to select useful rules from tree-ensembles, and rule selection is automated through a declarative encoding. The generated rule sets therefore not only act as interpretations for tree-ensemble models but are also explainable. 

We then evaluate our approach from two perspectives: the number and relevance of rules in the rule sets. The number of rules is often associated with interpretability, with a large number of rules being less desirable. Performance metrics such as classification accuracy, precision and recall can be used as a measure of relevance of the rules to the prediction task.

This paper makes the following contributions:
\begin{itemize}
    \item We present a novel application of Answer Set Programming (ASP) for interpreting machine learning models. We propose a method to generate explainable rule sets from tree-ensemble models with ASP. More generally, this work contributes to the growing body of knowledge on integrating symbolic reasoning with machine learning.
    \item We present how the rule set generation problem can be reformulated as an optimization problem, where we leverage existing knowledge on declarative pattern mining with ASP.
    \item To demonstrate the practical applicability of our approach, we provide both qualitative and quantitative results from evaluations with public datasets, where machine learning models are used in a realistic setting.
\end{itemize}

The rest of this paper is organized as follows. In Section 2 we review and discuss related works. In Section 3, we review tree-ensembles, ASP and pattern mining. Section 4 presents our method to generate rule sets from tree-ensembles using pattern mining and optimization encoded in ASP. Section 5 presents experimental results on public datasets. Finally in Section 6 we present the conclusions.

\section{Related Works}

Summarizing tree-ensembles has been studied in literature, see for example, Born Again Trees \cite{breimanBornAgainTrees1996}, defragTrees \cite{haraMakingTreeEnsembles2018} and inTrees \cite{dengInterpretingTreeEnsembles2019}. While exact methods and implementations differ among these examples, a popular approach to tree-ensemble simplification is to create a simplified decision tree model that approximates the behavior of the original tree-ensemble model. Depending on how the approximate tree model is constructed, this could lead to a deeper tree with an increased number of conditions which makes them difficult to interpret.

Integrating association rule mining and classification is also known, e.g., Class Association Rules (CARs)\cite{liuIntegratingClassificationAssociation1998}, where association rules discovered by pattern mining algorithms are combined to form a classifier. Repeated Incremental Pruning to Produce Error Reduction (RIPPER)\cite{COHEN1995115} was proposed as an efficient approach for classification based on association rule mining, and it is a well-known rule-based classifier. In CARs and RIPPER, rules are mined from data with dedicated association rule mining algorithms, then processed to produce a final classifier.


Interpretable classification models is another area of active research.  Interpretable Decision Sets (IDS)\cite{lakkarajuInterpretableDecisionSets2016} are learned through an objective function which simultaneously optimizes accuracy and interpretability of the rules. In Scalable Bayesian Rule Lists (SBRL)\cite{yangScalableBayesianRule2017}, probabilistic IF-THEN rule lists are constructed by maximizing the posterior distribution of rule lists. In RuleFit\cite{friedmanPredictiveLearningRule2008}, a sparse linear model is trained over rules extracted from tree-ensembles. RuleFit is the closest to our work in this regard, in the sense that both RuleFit and our method extract conditions and rules from tree-ensembles, but differ in the treatment of rules and representation of final rule sets. In RuleFit, rules are accompanied by regression coefficients, and it is left up to the user to further interpret the result.

Lundberg et al.\cite{lundbergLocalExplanationsGlobal2020} showed how a variant of SHAP\cite{lundberg2017unified}, which is a post-hoc interpretation method, can be applied to tree-ensembles. While our method does not produce importance measures for each feature, the information about which rule fired to reach the prediction can be offered as an explanation in a human readable format. Shakerin and Gupta \cite{shakerinInductionNonMonotonicLogic2019} proposed a method to use LIME weights\cite{ribeiroWhyShouldTrust2016} as a part of learning heuristics in inductive learning of default theories. Instead of learning rules with heuristics from data, our method directly handles rules which exist in decision tree models with answer set solver.


Guns et al.\cite{gunsItemsetMiningConstraint2011a} applied constraint programming (CP), a declarative approach, to itemset mining. This constraint satisfaction perspective led to the development of ASP encoding of pattern mining e.g.,  \cite{jarvisaloItemsetMiningChallenge2011a,guyetUsingAnswerSet}. Gebser et al.\cite{DBLP:conf/ijcai/GebserGQ0S16} applied preference handling to sequential pattern mining, and Paramonov et al.\cite{paramonovHybridASPbasedApproach2019} extended the declarative pattern mining by incorporating dominance programming (DP) from Negrevergne et al.\cite{negrevergneDominanceProgrammingItemset2013} to the specification of global constraints. Paramonov et al.\cite{paramonovHybridASPbasedApproach2019} proposed a hybrid approach where the solutions are effectively screened first with dedicated algorithms for pattern mining tasks, then declarative ASP encoding is used to extract condensed patterns. While aforementioned works focused on extracting interesting patterns from transaction or sequence data, our focus in this paper is to generate rule sets from tree-ensemble models to help users interpret the behavior of machine learning models. In terms of ASP encoding, we use dominance relations similar to the ones presented in Paramonov et al.\cite{paramonovHybridASPbasedApproach2019} to further constrain the search space.


\section{Background}

\subsection{Tree-Ensembles}

{\it Tree-Ensemble (TE)} models are machine learning models widely used in practice, typically, but not limited to, when learning from tabular datasets. A TE consist of multiple base decision trees each trained on an independent subset of the input data. For example, Random Forests \cite{breimanRandomForests2001} and Gradient Boosted Decision Tree (GBDT) \cite{friedmanGreedyFunctionApproximation2001} are tree-ensemble models. Recent surge of efficient and effective GBDT algorithms, e.g., LightGBM \cite{keLightGBMHighlyEfficient2017}, has led to wide adoption of TE models in practice. Although individual decision trees are considered to be interpretable \cite{huysmansEmpiricalEvaluationComprehensibility2011}, ensembles of decision trees are seen as less interpretable. 

The purpose of using TE models is to predict the unknown value of an attribute \(y\) in the dataset, referred to as \textit{labels}, using the known values of other attributes \(\mathbf{x}=(x_1,x_2,...,x_m)\), referred to as \(\textit{features}\). For brevity we restrict our discussion to classification problems. During the training or learning phase, each input instance to the TE models is a pair of features and labels, i.e. \((\mathbf{x}_i, y_i)\), where \(i\) denotes the instance index, and during the prediction phase, each input instance only include features, \((\mathbf{x}_i)\), and the model is tasked to produce predictions \(\hat{y}_i\). A collection of input instances, complete with features and labels, is referred to as a \textit{dataset}. 
Given a dataset \(\mathcal{D}=\{(\mathbf{x}_i, y_i)\}\) with \(n\in\mathbb{N}\) examples and \(m\in\mathbb{N}\) features, a decision tree classifier \(t\) will predict the class label \(\hat{y}_i\) based on the feature vector \(\mathbf{x}_i\) of the \(i\)-th sample: \(\hat{y}_i = t(\mathbf{x}_i)\). A tree-ensemble \(\mathcal{T}\) uses \(K\in\mathbb{N}\) trees and additionally an aggregation function \(f\) over the \(K\) trees which combines the output from the trees: \(\hat{y}_i = f(t_{k\in K}(\mathbf{x}_i))\). In the case of Random Forest, for example, \(f\) is a majority voting scheme (i.e. \texttt{argmax} of \texttt{sum}), and in GBDT \(f\) may be a summation followed by softmax to obtain \(\hat{y}_i\) in terms of probabilities.

In this paper a decision tree is assumed to be a binary tree where the internal nodes hold split conditions (e.g., \(x_1 \leq 0.5\)) and leaf nodes hold information related to class labels such as the number of supporting data points per class label that have been assigned to the leaf nodes.
Richer collections of decision trees provide higher performance and less uncertainty in prediction compared to a single decision tree. Typically, each TE model has specific algorithms for learning base decision trees, adding more trees and combining outputs from the base trees to produce the final prediction. In GBDT, the base trees are trained sequentially by fitting the residual errors from the previous step. Interested readers are referred to \cite{friedmanGreedyFunctionApproximation2001}, and its more recent implementations LightGBM \cite{keLightGBMHighlyEfficient2017} and XGBoost \cite{chenXGBoostScalableTree2016}.

\subsection{Answer Set Programming}
\textit{Answer Set Programming} \cite{lifschitzWhatAnswerSet2008} has its roots in logic programming and non-monotonic reasoning. A {\it normal logic program} is a set of rules of the form
\[\mathrm{a_1} \ \text{:-} \ \ \mathrm{a_2},\ \dots, \ \mathrm{a_m}, \ \mathrm{not} \ \mathrm{a_{m+1}}, \ \dots, \ \mathrm{not} \  \mathrm{a_n}.\]
where each \(\mathrm{a_i}\) is a first-order atom with \(1 \leq \mathrm{i} \leq \mathrm{n}\) and \texttt{not} is {\it default negation}. If only \(\mathrm{a_1}\) is included (\(\mathrm{n} = 1\)), the above rule is called a {\it fact}, whereas if \(\mathrm{a_1}\) is omitted, it represents an {\it integrity constraint}. A normal logic program induces a collection of models, which are called \textit{answer sets} defined by the stable model semantics \cite{gelfondStableModelSemantics1988}. Additionally, in modern ASP systems, constructs such as {\it conditional literals} and {\it cardinality constraints} are supported. The former in \textit{clingo} \cite{DBLP:journals/corr/GebserKKS14} are written in the form \(\{ \mathrm{a}(\texttt{X}) \ \text{:} \ \mathrm{b}(\texttt{X}) \}\)\footnote{Unless otherwise noted, we follow the Prolog-style notation in logic programs where strings beginning with a capital letter are variables, and others are predicate symbols or constants.}, and expanded into the conjunction of all instances of \(\mathrm{a}(\texttt{X})\) where corresponding \(\mathrm{b}(\texttt{X})\) holds. The latter are written in the form \(s_1 \ \{ \mathrm{a}(\texttt{X}) \ \text{:} \ \mathrm{b}(\texttt{X}) \} \ s_2 \), which is interpreted as \(s_1 \leq \texttt{\#count} \{ \mathrm{a}(\texttt{X}) \ \text{:} \ \mathrm{b}(\texttt{X}) \} \leq s_2 \) where \(s_1\) and \(s_2\) are treated as lower and upper bounds, respectively, thus the statement holds when the count of instances \(\mathrm{a}(\texttt{X})\) where \(\mathrm{b}(\texttt{X})\) holds, is between \(s_1\) and \(s_2\). The minimization (or maximization) of an objective function can be expressed with \(\texttt{\#minimize}\) (or  \(\texttt{\#maximize}\)). \textit{clingo} supports multiple optimization statements in a single program, and one can implement multi-objective optimization with priorities by defining two or more optimization statements.


\subsection{Pattern Mining}
In a general setting, the goal of pattern mining is to find interesting patterns from data, where patterns can be, for example, itemsets, sequences and graphs. For example, in \textit{frequent itemset mining} \cite{agrawalFastAlgorithmsMining1994}, the task is to find all subsets of items that occur together more than the threshold count in databases. In this work, the patterns of interest are sets of predictive rules. A \textit{predictive rule} has the form \( c \Leftarrow s_1 \wedge s_2 \wedge , ..., s_n \), where \(c\) is a class label, and \(\{s_i\}\) (\(1 \leq i \leq n\)) represents conditions. 

For pattern mining with constraints, the notion of \textit{dominance} is important, which intuitively reflects pairwise preference relation \((<^*)\) between patterns \cite{negrevergneDominanceProgrammingItemset2013}. Let \(C\) be a constraint function that maps a pattern to \(\{\top, \bot\}\), and let \(p\) be a pattern, then the pattern \(p\) is \textit{valid} iff  \(C(p)=\top\), otherwise it is \textit{invalid}. An example of \(C\) is a function that checks the support of a pattern is above the threshold. The pattern \(p\) is said to be \textit{dominated} iff there exists a pattern \(q\) such that \(p <^* q\) and \(q\) is valid under \(C\). Dominance relations have been used in ASP encoding for pattern mining \cite{paramonovHybridASPbasedApproach2019}.

There are existing ASP encodings of pattern mining algorithms, e.g., \cite{jarvisaloItemsetMiningChallenge2011a,DBLP:conf/ijcai/GebserGQ0S16,paramonovHybridASPbasedApproach2019}, that can be used to mine itemsets and sequences. Here we develop and apply our own encoding on rules to extract interesting rules from tree-ensembles. On the surface, our problem setting may appear similar to frequent itemset and sequence mining, however, rule set generation is different from these pattern mining problems. We can indeed borrow some ideas from frequent itemset mining for encoding, however, our goal is not to decompose rules (cf. transactions) into individual conditions (cf. items) then constructing rule sets (cf. itemsets) from conditions, but rather to treat each rule in its entirety then combining rules to form rule sets. The body (antecedent) of a rule can also be seen as a sequence, where the conditions are connected by conjunction connective \(\wedge\), however, in our case, the ordering of conditions does not matter, thus sequential mining encodings that use slots to represent positional constraints \cite{DBLP:conf/ijcai/GebserGQ0S16} cannot be applied directly to our problem. 


\section{Rule Set Generation}

\subsection{Problem Statement}
The rule set generation problem is represented as a tuple \(P=\{R,M,C,O\}\), where \(R\) is the set of all rules in the tree-ensemble, \(M\) is set of meta-data and properties associated with each rule in \(R\), \(C\) is the set of user-defined constraints including preferences, and \(O\) is the set of optimization objectives. The goal is to generate a set of rules from \(R\) by selection under constraints \(C\) and optimization objectives \(O\), where constraints and optimization may refer to the meta-data \(M\). In the following sections, we describe how we construct each \(R\), \(M\), \(C\) and \(O\), and finally how we solve this problem with ASP.

\subsection{Rule Extraction from Decision Trees}\label{sec:ruleextraction}
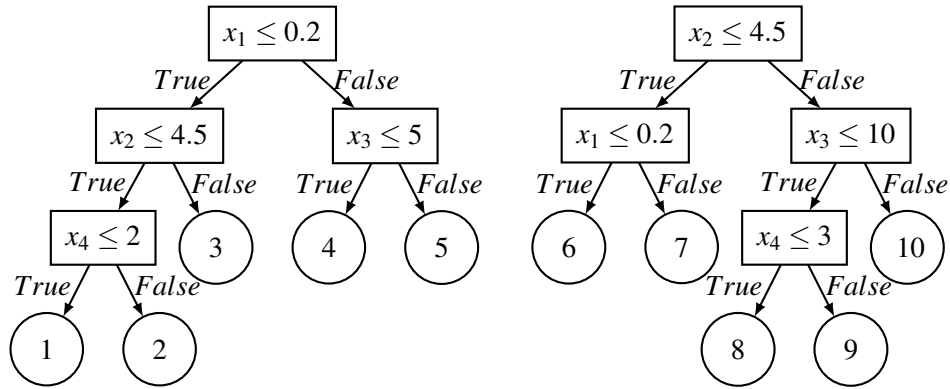
\begin{figure}[tb]
  \centering 

\begin{tikzpicture}

\tikzstyle{level 1}=[level distance=1cm, sibling distance=3cm]
\tikzstyle{level 2}=[level distance=1cm, sibling distance=1.5cm]
\tikzstyle{level 3}=[level distance=1cm, sibling distance=1.5cm]
\tikzstyle{level 4}=[level distance=1cm, sibling distance=1cm]

\tikzset{decision/.style={
draw=black, solid, rectangle, 
anchor=north,
inner sep=2mm, outer sep=0, text centered, growth parent anchor=south} 
}
\tikzset{prediction/.style={
draw=black, solid, circle, 
anchor=north,
text width=.3cm,
inner sep=2mm, outer sep=0, text centered, growth parent anchor=south} 
}
\tikzset{arrow/.style={-latex, thick}}
\tikzset{arrow-dashed/.style={-latex, dashed, thick}}

\node [decision] (t1) {$ x_1 \leq 0.2 $} [solid, -latex, thick]
  child { 
    node [decision] {$ x_2 \leq 4.5$}
      child {
        node [decision] {$x_4 \leq 2$} [solid, -latex, thick]
          child {
            node [prediction] {1} [solid, -latex, thick]
          edge from parent node [left,above,xshift=-0.45cm,yshift=-0.2cm] {$ True $}}
          child {
            node [prediction] {2} [solid, -latex, thick]
          edge from parent node [right,above,xshift=0.5cm,yshift=-0.2cm] {$ False $}}
      edge from parent node [left,above,xshift=-0.45cm,yshift=-0.2cm] {$ True $} }
      child {
        node [prediction] {3} [solid, -latex, thick]
      edge from parent node [right,above,xshift=+0.5cm,yshift=-0.2cm] {$ False $}}
  edge from parent node [left,above,xshift=-0.45cm,yshift=-0.2cm] {$ True $} } 
  child {
    node [decision] {$x_3 \leq 5$}
      child {
        node [prediction] {4}
      edge from parent node [left,above,xshift=-0.45cm,yshift=-0.2cm] {$ True $}}
      child {
        node [prediction] {5} [solid, -latex, thick]
      edge from parent node [right,above,xshift=+0.5cm,yshift=-0.2cm] {$ False $}}
  edge from parent node [right,above,xshift=+0.5cm,yshift=-0.2cm] {$ False $}}
;

\node [decision, right=4.5cm of t1] {$ x_2 \leq 4.5 $} [solid, -latex, thick]
  child { 
    node [decision] {$ x_1 \leq 0.2$}
      child {
        node [prediction] {6}
        edge from parent node [right,above,xshift=-0.45cm,yshift=-0.2cm] {$ True $}
        }
      child {
        node [prediction] {7} [solid, -latex, thick]
      edge from parent node [right,above,xshift=+0.5cm,yshift=-0.2cm] {$ False $}}
  edge from parent node [left,above,xshift=-0.45cm,yshift=-0.2cm] {$ True $} } 
  child {
    node [decision] {$x_3 \leq 10$}
        child {
        node [decision] {$x_4 \leq 3$} [solid, -latex, thick]
          child {
            node [prediction] {8} [solid, -latex, thick]
          edge from parent node [left,above,xshift=-0.45cm,yshift=-0.2cm] {$ True $}}
          child {
            node [prediction] {9} [solid, -latex, thick]
            edge from parent node [right,above,xshift=0.5cm,yshift=-0.2cm] {$ False $}}
          edge from parent node [left,above,xshift=-0.45cm,yshift=-0.2cm] {$ True $}}
      child {
        node [prediction] {10} [solid, -latex, thick]
      edge from parent node [right,above,xshift=+0.5cm,yshift=-0.2cm] {$ False $}}
  edge from parent node [right,above,xshift=+0.5cm,yshift=-0.2cm] {$ False $}}
;
\end{tikzpicture}
    \vspace{-.5\baselineskip}
  \caption{A simple decision tree-ensemble consisting of two decision trees. The rule associated with each node is given by the conjunction of all conditions associated with nodes on the paths from the root node to that node.}
  \label{fig:simpletree}
  \vspace{-.5\baselineskip}
\end{figure}

Recall that a tree-ensemble \(\mathcal{T}\) is a collection of \(K\) decision trees, and we refer to individual trees \(t_k\) with subscript \(k\). An example of a decision tree-ensemble is shown in Figure \ref{fig:simpletree}. A decision tree \(t_k\) has \(N_{t_k}\) nodes and \(L_{t_k}\) leaves. Each node represents a split condition and there are \(L_{t_k}\) paths from the root node to the leaves. For simplicity, we assume only features that have orderable values (continuous features) are present in the dataset in the examples below.\footnote{Real datasets may have unorderable categorical values. For example, in the \textit{census} dataset, occupation (Sale, etc.) and education (Bachelors, etc.) are categorical features. Support for categorical feature split is implementation-dependent, however in general one can replace the continuous split with a subset selection e.g., \(x_c \in \{x_{c1}, x_{c2},...\}\)} The tree on the left in Figure \ref{fig:simpletree} has 4 internal nodes including the root node with condition \([x_1 \leq 0.2]\) and 5 leaf nodes, therefore there are 5 paths from the root note to the leaf nodes 1 to 5. 

From the left-most path of the decision tree on the left in Figure \ref{fig:simpletree}, the following prediction rule is created. We assume that node 1 predicts class label 1 in this instance.\footnote{Label=1 and 0 refer to the attributes in the dataset and have different meaning depending on the dataset. For example, in the \textit{census} dataset, label=1 and 0 mean that the personal income is more than \$50,000 and that it is no more than \$50,000, respectively.}
\begin{equation*}
    class(1) \Leftarrow (x_1 \leq 0.2) \wedge (x_2 \leq 4.5) \wedge (x_4 \leq 2)
\end{equation*}
Assuming that node 2 predicts class label 0, we also construct the following rule (note the reversal of the condition on \(x_4\)):
\begin{equation*}
    class(0) \Leftarrow (x_1 \leq 0.2) \wedge (x_2 \leq 4.5) \wedge (x_4 > 2)
\end{equation*}
We can also construct subsets of rules by applying each of the conditions sequentially and computing the predicted label at each step. For example, from the last rule we may construct the following rule:
\begin{equation*}
    class(1) \Leftarrow (x_1 \leq 0.2) \wedge (x_2 \leq 4.5)
\end{equation*}

The set of all rules, \(R\), is constructed as follows: 
\begin{enumerate}
    \item Enumerate all possible paths from the root node to the leaves. For a binary decision tree with depth \(d_k\), the maximum number of leaf nodes is \(2^{d_k}\), which is also the maximum number of paths from the root node to the leaf nodes.
    \item For each path, at each subsequent node on the path to the leaf node, the split condition of the node is appended to the body (antecedent, set of conditions) of the rule. For a decision tree the maximum number of such rules is the same as the maximum number of nodes in the tree, i.e. \(2^{d_k + 1}-1\).
    \item Compute the predicted class label for each rule. For simplicity, we apply all conditions in the rule and calculate the most likely class label from the count data (\(\texttt{argmax}\) of counts).
    \item Add the generated rules to the candidate rule set \(R\).
    \item Repeat steps 1 to 4 for each tree \(t_k\) where \(1 \leq k \leq K\), in the ensemble of \(K\) trees.
\end{enumerate}

By constructing the candidate rule set \(R\) in this way, the bodies (antecedents) of rules included in rule sets are guaranteed to exist in at least one of the trees in the tree ensemble. Rule sets generated in this manner are therefore faithful to the representation of the original model in this sense. If we were to construct rules from the unique set of split conditions, the resulting rule may have combinations of conditions that may not exist in any of the trees.


\subsection{Computing Metrics and Meta-data for Selection}\label{sec:metrics}
After the candidate rule set \(R\) is constructed, we gather information about the performance and properties of each rule and collect them into a set \(M\). Performance metrics, in general, measure how well a rule can predict class labels. Examples of widely adopted performance metrics in machine learning are: accuracy, precision, recall and F1-score. We compute multiple metrics for a single rule, to meet a range of user requirements for interpretation. For example, one user may only be interested in simply most accurate rules (maximize accuracy), whereas another user could be interested in more precise rules (maximize precision), or rules with more balanced performance (maximize F1-score). The meta-data, or properties, of a rule are information such as the size of the rule, as defined by the number of conditions in the rule, or the number of instances which are covered by the rule. These properties can be used in the selection step to define competing objectives. For example, one can expect a very long rule with relatively large number of rules to be precise, but the rule may be too specific and may not cover many instances. Moreover, a long rule is more difficult to comprehend than a short, concise rule. In this case, the size property needs to be minimized, while the precision metric is maximized.

The candidate rule set \(R\) and meta-data set \(M\) are represented as facts in ASP, as shown in Table \ref{tab:atomdescription}. For example, the first rule in Section \ref{sec:ruleextraction} may be represented as follows\footnote{The performance metrics are for illustration purposes only and are chosen arbitrarily.}:

\begin{Verbatim}[frame=single,fontsize=\small]
% rule 1
rule(1). condition(1,1). condition(1,2). condition(1,3). support(1,10).
size(1,3). accuracy(1,50). error_rate(1,50). precision(1,30). 
recall(1,40). f1_score(1,34). predict_class(1,1).
\end{Verbatim}

\begin{table}
    \caption{List of predicates representing a rule in ASP.}
    \label{tab:atomdescription}
    \begin{tabular}{ll}
    \hline\hline
    Predicate & Meaning \\
    \hline
    \small{\texttt{rule(X)}}              & \texttt{X} holds the rule index. \\
    \small{\texttt{condition(X,I)}}       & Rule \texttt{X} has condition \texttt{I}. \\
    \small{\texttt{support(X,S)}}         & Support \texttt{S} of rule \texttt{X}, the number of instances that is covered by rule \texttt{X}.\\
    \small{\texttt{size(X,L)}}            & Number of conditions in rule \texttt{X} (length, \texttt{L}). \\
    \small{\texttt{error\_rate(X,E)}}     & Error rate (\(1 - accuracy\)), \texttt{E}, of the rule \texttt{X} evaluated in the training data. \\
    \small{\texttt{accuracy(X,A)}}        & Accuracy score of rule \texttt{X}. \\
    \small{\texttt{precision(X,P)}}       & Precision score of rule \texttt{X}. \\
    \small{\texttt{recall(X,R)}}          & Recall score of rule \texttt{X}. \\
    \small{\texttt{f1\_score(X,F)}}       & F1-score of rule \texttt{X}. \\
    \small{\texttt{predict\_class(X,C)}}  & Predicted class label \texttt{C} of rule \texttt{X}.\\
    \hline\hline
    \end{tabular}
    \vspace{-.5\baselineskip}
\end{table}


\subsection{Encoding Constraints}\label{sec:encodingconstraints}
For the rule set generation task, we consider three types of constraints: (1) local constraints that are applied on a per-rule basis, for example, to select rules that meet the minimum support threshold, (2) pairwise constraints that are applied to pairs of rules, which include dominance relations, and (3) global constraints that are applied to a set of rules, for example to control the total number of conditions in the rule set.

To encode local constraints, a predicate \texttt{valid(X)} is introduced, to specify that a \texttt{rule(X)} is valid whenever \texttt{invalid(X)} cannot be inferred:
\begin{Verbatim}[frame=single,fontsize=\small]
valid(X) :- rule(X), not invalid(X).
\end{Verbatim}
This example of a local constraint eliminates rules with low support:
\begin{Verbatim}[frame=single,fontsize=\small]
% this will eliminate rules that apply to less than 10 instances
invalid(X) :- rule(X), support(X,S), S < 10.
\end{Verbatim}

Pairwise constraints can be used to encode dominance relations between rules. For a rule \texttt{X} to be dominated by \texttt{Y}, \texttt{Y} must be strictly better in one criterion than \texttt{X} and at least as good as \texttt{X} or better in other criteria. For example, in the following case we encode the dominance relation between rules using the F1 score, support and size of the rule, where we prefer rules that are small (more interpretable), have higher support (covers more instances) and perform well (higher F1 score).
\begin{Verbatim}[frame=single,fontsize=\small]
% cannot be dominated
:- dominated.
% X is dominated by Y if ...
ge_f1_leq_size_geq_sup(Y) :- selected(X), valid(Y), size(X,Sx), size(Y,Sy), 
    f1_score(X,Fx), f1_score(Y,Fy), support(X,Spx), support(Y,Spy),
    Fx < Fy, Sx >= Sy, Spx <= Spy.
geq_f1_le_size_geq_sup(Y) :- selected(X), valid(Y), size(X,Sx), size(Y,Sy),
    f1_score(X,Fx), f1_score(Y,Fy), support(X,Spx), support(Y,Spy),
    Fx <= Fy, Sx > Sy, Spx <= Spy.
geq_f1_leq_size_ge_sup(Y) :- selected(X), valid(Y), size(X,Sx), size(Y,Sy),
    f1_score(X,Fy), f1_score(Y,Fy), support(X,Spi), support(Y,Spy),
    Fx <= Fy, Sx >= Sy, Spx < Spy.
dominated :- valid(Y), ge_f1_leq_size_geq_sup(Y).
dominated :- valid(Y), geq_f1_le_size_geq_sup(Y).
dominated :- valid(Y), geq_f1_leq_size_ge_sup(Y).
\end{Verbatim}

Global constraints are applied to rule sets in addition to the local and pairwise constraints and preferences. For example, the following "generator" encoding puts a limit on the maximum size of rule sets that are considered:
\begin{Verbatim}[frame=single,fontsize=\small]
% pick at least 1 rule and at maximum 10 rules for each predict_class
1 { selected(X) :  predict_class(X, K), valid(X) } 10 :- class(K).
\end{Verbatim}
This encoding will select at least 1 and up to 10 valid rules for each class label \texttt{K}. The properties of rule sets can also be used to construct constraints. For instance, one can put restrictions the maximum number of conditions in rule sets, using the aggregate atom \texttt{\#sum}:
\begin{Verbatim}[frame=single,fontsize=\small]
% total number of conditions should not exceed 30
:- #sum { S,X : size(X,S), selected(X) } > 30.
\end{Verbatim}
Exact set of constraints and preferences depend on the problem domain, use-case and/or intention of the user. The expressiveness of the ASP language allows one to represent constraints in a declarative manner under the semantics of logic programming.

\subsection{Optimizing Rule Sets}\label{sec:optimizingrulesets}
Finally, we pose the rule set generation problem as a multi-objective optimization problem, given aforementioned facts and constraints encoded in ASP. The desiderata for generated rule sets may contain multiple competing objectives. For instance, we consider a case where the user wishes to collect accurate rules that cover a large number of instances, while minimizing the number of conditions in the set. This is encoded as a group of optimization statements:
\begin{Verbatim}[frame=single,fontsize=\small]
% maximize accuracy and support, minimize the number of conditions
#maximize { A,X : selected(X), accuracy(X,A)}.
#maximize { S,X : selected(X), support(X,S)}.
#minimize { L,X : selected(X), size(X,L)}.
\end{Verbatim}

For optimization, we introduce a measure of overlap between the rules to be minimized. Intuitively, minimizing this objective should result in rule sets where rules share only a small number of conditions, which should further improve the interpretability of the resulting rule sets. Specifically, we introduce a predicate \texttt{rule\_overlap(X,Y,Cn)} to measure the degree of overlap between rules \texttt{X} and \texttt{Y}.
\begin{Verbatim}[frame=single,fontsize=\small]
% number of shared conditions between rules
rule_overlap(X,Y,Cn) :- selected(X), selected(Y), X!=Y,
    Cn = #count { Cx : Cx=Cy, condition(X,Cx), condition(Y,Cy) }.
#minimize { Cn,X : selected(X), selected(Y), rule_overlap(X,Y,Cn) }.
\end{Verbatim}

\section{Experiments}
We evaluate our rule set generation framework on several public datasets and compare the performance to existing methods including rule-based classifiers.

\subsection{Experimental Setup}
We used 10 publicly available datasets from the UCI Machine Learning Repository\footnote{\url{https://archive.ics.uci.edu/ml/index.php}} \cite{Dua:2019}. The summary of these datasets is shown in Table \ref{tab:datadescription}.
We used \textit{Clingo} 5.4.0\footnote{\url{https://potassco.org/clingo/}} \cite{DBLP:journals/corr/GebserKKS14} for answer set programming, and set the time out to 600 seconds.\footnote{Full ASP encoding of our method is available in the supplementary materials.} We used RIPPER implemented in Weka \cite{wittenWEKAWorkbenchOnline2016} and an open source implementation of RuleFit\footnote{\url{https://github.com/christophM/rulefit}} where Random Forest was selected as the rule generator, and scikit-learn\footnote{\url{https://scikit-learn.org/}} \cite{scikit-learn} for general machine learning functionalities.
Our experimental environment is a desktop machine with Ubuntu 18.04, Intel Core i9-9900K 3.6GHz (8 cores/16 threads) and 64GB RAM. 

\begin{table}
    \centering
    \caption{Datasets used in the experiments. \textit{\#data} and \textit{\#feature} refer to the number of data points (rows) and features (columns), respectively. The number of categorical features is shown in parenthesis.}
    \label{tab:datadescription}
    \begin{tabular}{lrrl}
    \hline\hline
        Dataset & \# \textit{data} & \# \textit{feature} & \(y\) = 1 \\
    \hline
        autism              & 704   & 20 (18)   & screening result \\
        breast              & 699   & 9 (9)     & malignant \\
        census              & 299,286 & 42 (29) & income \(>\) 50k \\
        credit\_a           & 690   & 14 (8)    & application accepted \\
        credit\_t           & 30,000 & 23 (10)  & payment next month \\
        heart               & 270   & 13 (8)    & disease present \\
        ionosphere          & 351   & 34 (0)    & good radar return \\
        kidney              & 400   & 24 (13)   & chronic disease \\
        krvskp              & 3,196  & 36 (36)  & white can win \\
        voting              & 435   & 16 (16)   & democrat \\
    \hline\hline
    \end{tabular}
    \vspace{-.5\baselineskip}
\end{table}

In order to evaluate the performance of the extracted rule sets, we implemented a naive rule-based classifier which is constructed from the rule sets extracted with our method. In this classifier, we apply the rules sequentially to the validation dataset and if all conditions within a rule are true for an instance in the dataset, the consequent of the rule is returned as the predicted class. More formally, given a set of rules \(R_s \subset R\) with cardinality \(|R_s|\) that shares the same consequent \(class(Q)\), we represent this rule-based classifier as the disjunction of antecedents of the rules:
\begin{equation*}
    class(Q) \Leftarrow body(R_1) \vee body(R_2) \vee ... \vee body(R_r) \textrm{ where } 1 \leq r \leq |R_s|
\end{equation*}
For a given data point, it is possible that there are no rules applicable, and in such cases the most common class label in the training dataset is returned.

We conduct the evaluation experiment in the following order. First, we train Random Forest and LightGBM on the datasets in Table \ref{tab:datadescription}.
We then apply our rule set generation method to the trained tree-ensemble models. Finally, we construct a naive rule-based classifier using the set of rules extracted in the previous step, and calculate performance metrics on the validation set. This process is repeated in a 5-fold stratified cross validation setting to estimate the performance. We compare the characteristics of our approach against the known methods RIPPER and RuleFit.

\begin{table}[htb]
    \centering
    \caption{Average number of candidate rules (\(|R|\)), size of the generated rule sets (\# rule), averaged over 5 folds. Hyphen indicates a failure case where no rules could be found.}
    \label{tab:numrules}
    \begin{tabular}{l | rr | rrr | r | r}
    \hline\hline
        & \multicolumn{2}{c|}{LightGBM+ASP} & \multicolumn{3}{c|}{RandomForest+ASP} & RuleFit & RIPPER \\
Dataset & \multicolumn{1}{c}{\(|R|\)} & \multicolumn{1}{c|}{\# rule} & & \multicolumn{1}{c}{\(|R|\)} & \multicolumn{1}{c|}{\# rule} & \multicolumn{1}{c|}{\# rule} & \multicolumn{1}{c}{\# rule} \\
\hline
autism & 2.0 & \textbf{1.0} & & 59.8 & 7.6 & 3.0 & 2.0 \\
breast & 131.2 & \textbf{2.8} & &  27.8 & 8.8 & 55.8 & 13.0 \\
census & 8806.8 & \textbf{9.0} &  & - & - & 304.0 & 54.7 \\
credit\_a & 275.2 & \textbf{3.8} &  & 123.4 & 7.4 & 55.2 & 7.0 \\
credit\_t & 2098.4 & \textbf{6.6} &  & - & - & 187.8 & 7.4 \\
heart & 159.6 & \textbf{2.8} &  & 47.6 & 8.8 & 40.8 & 6.2 \\
ionosphere & 314.4 & \textbf{5.2} &  & 1127.0 & 9.8 & 272.0 & 7.0 \\
kidney & 179.6 & \textbf{3.2} &  & 101.0 & 5.8 & 160.6 & 4.4 \\
krvskp & 140.8 & \textbf{7.6} &  & 69.6 & 10.0 & 240.4 & 16.4 \\
voting & 59.6 & \textbf{1.4} &  & 45.2 & 3.4 & 44.0 & 6.2 \\
    \hline\hline
    \end{tabular}
    \vspace{-.5\baselineskip}
\end{table}

\begin{table}[htb]
    \centering
    \caption{Average ratio of rule-based classifier's performance vs. original tree-ensembles. Acc.=accuracy, Prec.=Precision, Rec.=Recall and F1=F1 score. Performance ratio of 1 means the rule set's performance is identical to the original classifier. Hyphen indicates a failure case where no rules could be found.}
    \label{tab:performanceratio}
    \begin{tabular}{l | rrrr | rrrr | rrrr }
    \hline\hline
        & \multicolumn{4}{c|}{LightGBM+ASP} & \multicolumn{4}{c|}{RandomForest+ASP} & \multicolumn{4}{c}{RuleFit}\\
    Dataset  & Acc. & Prec. & Rec. & F1 & Acc. & Prec. & Rec. & F1 & Acc. & Prec. & Rec. & F1\\
    \hline
autism & \textbf{1.00} & \textbf{1.00} & \textbf{1.00} & \textbf{1.00} & 0.70 & 0.47 & \textbf{1.20} & 0.69 & 1.05 & 1.00 & \textbf{1.21} & 1.11\\
breast & 0.75 & 0.62 & \textbf{1.05} & 0.77 & 0.76 & 0.61 & \textbf{1.08} & 0.78 & 1.01 & 1.00 & \textbf{1.03} & 1.01\\
census & 0.37 & 0.12 & \textbf{2.01} & 0.27 & - & - & - & - & - & - & - & -\\
credit\_a & 0.81 & 0.78 & \textbf{0.99} & 0.85 & 0.94 & 0.89 & \textbf{1.05} & 0.96 & 1.02 & 0.97 & \textbf{1.10} & 1.03\\
credit\_t & 0.39 & 0.35 & \textbf{2.49} & 0.79 & - & - & - & - & - & - & - & -\\
heart & 0.83 & 0.79 & \textbf{0.99} & 0.85 & 0.69 & 0.59 & \textbf{1.40} & 0.86 & 1.04 & 0.98 & \textbf{1.17} & 1.08\\
ionosphere & 0.80 & 0.85 & \textbf{0.93} & 0.87 & 0.69 & 0.71 & \textbf{1.01} & 0.83 & 1.01 & \textbf{1.03} & 0.98 & 1.00\\
kidney & 0.74 & 0.73 & \textbf{0.99} & 0.83 & 0.63 & 0.64 & \textbf{1.00} & 0.78 & \textbf{1.01} & \textbf{1.01} & 1.00 & 1.00\\
krvskp & 0.78 & 0.73 & \textbf{0.93} & 0.82 & 0.58 & 0.60 & \textbf{1.03} & 0.75 & 1.09 & \textbf{1.14} & 1.02 & 1.08\\
voting & 0.94 & \textbf{0.95} & \textbf{0.95} & \textbf{0.95} & 0.66 & 0.64 & \textbf{1.08} & 0.81 & 1.03 & 1.01 & \textbf{1.04} & 1.02 \\
    \hline\hline
    \end{tabular}
\end{table}


\subsection{Number of Rules}

The average number of rules extracted from the data is shown in Table \ref{tab:numrules}. RuleFit includes original features (called linear terms) as well as conditions extracted from the tree-ensembles in the construction of a sparse linear model, that is to say, the counts in Table \ref{tab:numrules} may be inflated by the linear terms. On the other hand, the output from RIPPER only contains rules, and RIPPER has rule pruning and rule set optimization to further reduce the rule set size. Moreover, RIPPER has direct control over which conditions to include into rules, whereas our method and RuleFit relies on the structure of the decision trees to construct rules. 

Our approach consistently produces smaller rule sets compared to RuleFit, and the rule sets are comparable in size to, or smaller than, those produced by RIPPER. Comparing the size of the candidate rule set \(|R|\) with the size of rule sets, our method can produce rule sets which are significantly smaller than the original model. Overall, in terms of the number of rules in the final rule set, where smaller count is desirable for better interpretability, LightGBM+ASP performed the best, followed by RIPPER. The failure cases with Random Forest (\textit{census} and \textit{credit\_t} datasets) occurred due to leaf-only trees. Because leaf-only trees have no split conditions, rules could not be extracted and our method produced no rule sets as a result.

\subsection{Relevance of Rules}

To quantify the relevance of the extracted rules, we measured the ratio of performance metrics using the naive rule-based classifier by 5-fold cross validation (Table \ref{tab:performanceratio}). Performance ratio of less than 1.0 means that the rule-based classifier performed worse than the original classifier (LightGBM and Random Forest), whereas performance ratio greater than 1 means the rule set's performance is better than the original classifier.

From Table \ref{tab:performanceratio} we observe that this particular encoding yields rules that have good recall, but other metrics could suffer especially in larger datasets such as \textit{census} and \textit{credit\_t}. In this instance, F1-score was used to define dominance relations in the ASP encoding, and the performance is mostly comparable with the original model, with the exception of the \textit{census} dataset where the F1-score was noticeably worse. For this evaluation, we did not set any restrictions on the number of rules RuleFit could have, and it performs as well as the original Random Forest classifier in most cases.



    

\subsection{Changing Optimization Criteria}

The definition of optimization objectives has a direct influence over the performance of the resulting rule sets, and the objectives need to be set in accordance with user requirements. Because the solution space is bound by the constraints, changing the optimization statements by themselves may not give desired solutions. In an extreme case, e.g., LightGBM+ASP on the \textit{autism} dataset, there is only 1 candidate rule to begin with and changing the optimization statements (e.g., more weight on precision) will have no effect on the final solution.

The answer sets found by \textit{clingo} with multiple optimization statements are optimal with respect to the set of goals defined by the user. Instead of using accuracy one may use other rule metrics as defined in Table \ref{tab:atomdescription} such as precision and/or recall. If there are priorities between optimization criteria, then one could use the priority notation (\texttt{weight@priority}) in \textit{clingo} to define them. 
 Optimal answer sets can be computed in this way, however, if enumeration of such optimal sets is important, then one could use the \texttt{pareto} or \texttt{lexico} preference definitions provided by \textit{asprin} \cite{brewkaAsprinCustomizingAnswer2015} to enumerate Pareto optimal answer sets. Instead of presenting a single optimal rule set to the user, this will allow the user to explore other optimal rule sets.

\section{Conclusion}
In this work, we presented a method for generating explainable rule sets from tree-ensembles using pattern mining techniques encoded in ASP for the interpretation of tree-ensembles. 
Adopting the declarative programming paradigm with ASP allows the user to take advantage of the expressiveness of ASP in representing constraints and preferences. This makes our approach particularly suitable for situations where fast  prototyping is required, since changing the constraint and preference settings require relatively low effort compared specialized mining algorithms. Useful interpretations can be generated using our approach, and combined with the expressive ASP encoding, we hope that our method will help the users of tree-ensemble models to better understand the behavior of such models.

A limitation of our method in terms of scalability is the size of search space, which is exponential in the number of valid rules.
When the number of candidate rules is large, we suggest using stricter local constraints on the rules, or reducing the maximum number of rules to be included into rule sets (Section \ref{sec:encodingconstraints}), in order to achieve reasonable solving time.



There is a number of directions for further research. First, while the current work did not modify the conditions in the rules in any way, rule simplification approaches could be incorporated to remove redundant conditions. Second, we could extend the current work to support regression problems. More generally, in future, we plan to explore how ASP and modern statistical machine learning could be integrated effectively to produce more interpretable machine learning systems.

\section*{Acknowledgments}
This work has been supported by JSPS KAKENHI Grant No. 21H04905.

\bibliographystyle{eptcs}
\bibliography{ref}

\end{document}